\documentclass[review]{elsarticle}

\usepackage{lineno,hyperref}
\modulolinenumbers[5]

\journal{Neurocomputing}

\usepackage[utf8]{inputenc} 
\usepackage{dsfont} 
\usepackage{amsmath} 
\usepackage[ruled,vlined,linesnumbered]{algorithm2e} 
\usepackage{graphicx} 
\graphicspath{{./images/}}
\DeclareGraphicsExtensions{.pdf,.jpeg,.png}
\usepackage[caption=false,font=footnotesize]{subfig} 
\usepackage{booktabs} 

\begin{document}

\begin{frontmatter}

\title{Particle Competition and Cooperation for Semi-Supervised Learning with Label Noise}


\author[unesp]{Fabricio A. Breve\corref{cor1}}
\ead{fabricio@rc.unesp.br}

\author[usp]{Liang Zhao}
\ead{zhao@usp.br}

\author[unifesp]{Marcos G. Quiles}
\ead{quiles@unifesp.br}

\cortext[cor1]{Corresponding author}

\address[unesp]{Department of Statistics, Applied Mathematics and Computation (DEMAC), Institute of Geosciences and Exact Sciences (IGCE), São Paulo State University (UNESP), Avenida 24A, 1515 - DEMAC, Bela Vista, Rio Claro, São Paulo, CEP 13506-900, Brazil}

\address[usp]{Department of Computer Science and Mathematics (DCM), School of Philosophy, Science and Literature in Ribeirão Preto (FFCLRP), University of São Paulo (USP), Av. Bandeirantes, 3900, Monte Alegre, Ribeirão Preto, São Paulo, CEP 14040-900, Brazil}

\address[unifesp]{Institute of Science and Technology (ICT), Federal University of São Paulo (Unifesp), São José dos Campos, SP, Brazil}





\begin{abstract}
Semi-supervised learning methods are usually employed in the classification of data sets where only a small subset of the data items is labeled. In these scenarios, label noise is a crucial issue, since the noise may easily spread to a large portion or even the entire data set, leading to major degradation in classification accuracy. Therefore, the development of new techniques to reduce the nasty effects of label noise in semi-supervised learning is a vital issue. Recently, a graph-based semi-supervised learning approach based on Particle competition and cooperation was developed. In this model, particles walk in the graphs constructed from the data sets. Competition takes place among particles representing different class labels, while the cooperation occurs among particles with the same label. This paper presents a new particle competition and cooperation algorithm, specifically designed to increase the robustness to the presence of label noise, improving its label noise tolerance. Different from other methods, the proposed one does not require a separate technique to deal with label noise. It performs classification of unlabeled nodes and reclassification of the nodes affected by label noise in a unique process. Computer simulations show the classification accuracy of the proposed method when applied to some artificial and real-world data sets, in which we introduce increasing amounts of label noise. The classification accuracy is compared to those achieved by previous particle competition and cooperation algorithms and other representative graph-based semi-supervised learning methods using the same scenarios. Results show the effectiveness of the proposed method.
\end{abstract}


\end{frontmatter}


\section{Introduction}

Label noise is an important issue in machine learning and, more specifically, in data classification. A classifier usually learns from a set of labeled samples to predict the classes of new samples. However, many real-world data sets contain noise, and learning from those may lead to many potential negative consequences \cite{Frenay2013}. Label noise may be of two different types: feature noise and class noise \cite{Frenay2013,Zhu2004}. Feature noise affects observed values of the data features. For example, sensors may introduce some Gaussian noise during data feature measurement. On the other hand, class noise alters the labels assigned to data instances. For instance, a specialist may mistakenly assign the wrong class to some samples \cite{Hickey1996}, specially when the labeling task is subjective, like in medical applications \cite{Malossini2006}. In this paper, we focus on class noise, which is potentially the more harmful type of label noise \cite{Frenay2013,Zhu2004,Saez2014}.

The reliability of class labels is important in supervised learning algorithms \cite{Slonim1996,Krishnan1988}, but in semi-supervised learning this is a crucial issue. Semi-supervised learning is usually applied to problems where only a small subset of labeled samples is available, together with a large amount of unlabeled samples \cite{Zhu2005,Chapelle2006,Abney2008}. This is a common situation nowadays, as the size of the data sets being treated is constantly increasing, making prohibitive the task of labeling samples to supervised approaches. This task is time consuming and usually requires the work of human experts. Therefore, class noise is a major problem in semi-supervised learning, due to the smaller proportion of labeled data in the whole data set. In these scenarios, errors may easily affect the classification of a large portion or even the entire data set \cite{Breve2012SBRN}, leading to major degradation in classification accuracy, which is the more frequently reported consequence of label noise \cite{Frenay2013}. Therefore, it is vital to develop techniques to reduce the nasty effects of label noise in semi-supervised learning process.

There are three broader approaches of handling label noise in classification \cite{Teng2001,Frenay2013,Yin2011}: robust algorithms, filtering, and correction. Robust algorithms are designed to naturally tolerate a certain amount of label noise, so they do not need any special treatment. Filtering the noise means that some label noise cleaning strategy is used to identify and discard noisy labels before the training process. Finally, correction means that the noisy labels are identified, but instead of eliminating them, they are repaired or handled properly. Albeit it is not always clear whether an approach belongs to one category or the other. \cite{Frenay2013}. Usually, a mixed strategy of the above mentioned categories are used to deal with label noise problem.

Recently, a particle competition and cooperation approach was used to realize graph-based semi-supervised learning \cite{Breve2012TKDE}. The data set is converted into a graph, where samples are nodes with edges between the similar samples. Each labeled node is associated with a labeled particle. Particles walk through the graph and cooperate with identically labeled particles to classify unlabeled samples, while competing against particles with different labels. The main advantage of particle competition and cooperation method over most other semi-supervised learning algorithms can be summarized as follows: we have proved that it has lower computational complexity \cite{Breve2012TKDE} due to its local propagation nature; at the same time, extensive numerical studies show the method can achieve high precision of classification; it is similar to many natural or biological processes, such as resource competition by animals, territory exploration by humans (animal), election campaigns, etc. In this way, we believe that the particle competition and cooperation method can be also used back to model those natural or biological systems. The original competition and cooperation process generates much useful information and saved in the dominance level vector of each node. Such information can be used to solve other relevant problems beyond the standard machine learning tasks. For example, it can help to determine data class overlapping, fuzzy classification, and outlier detection by analyzing the distribution of the dominance vectors \cite{Breve2013SoftComputing}. In this paper, we modify and further improve the original method to treat an important issue in semi-supervised learning: learning with label noise or wrong labels.

Taking the interesting features of the particle competition and cooperation approach into account, further improvements to increase the robustness of the method have been pursued. Some preliminary results were presented in \cite{Breve2012SBRN}. The improved algorithm raised classification accuracy in the presence of label noise. However, some drawbacks have been identified, like high differences in node degree among labeled and unlabeled nodes and lack of connection between labeled particles and their corresponding labeled nodes. As a consequence, the particles spend quite more time on labeled nodes than unlabeled ones, which demands a higher number of iterations to converge. Moreover, on conditions where the amount of label noise is critical, a team of particles may switch territory with another team. This happens because particles are not strongly attracted to their corresponding nodes and they may be attracted to nodes with label noise which are on another class territory. This territory switching phenomenon always involves all particles from two or more classes, therefore it leads to major classification accuracy lost.

In this paper, we further improved the robustness of the particle competition and cooperation method to label noise. We addressed the problems of the preliminary version by enhancing graph generation, leveling nodes degrees, and thus lowering execution times. The territory switching phenomenon was also addressed by the changes in the graph generation, changes in the particles distance tables calculation, and periodic resets in particles and nodes. These improvements allow the new model to keep the particles closer to their neighborhood, increase the attraction between particles and their corresponding labeled nodes, and bring particles back after a while if they still fail to avoid territory switching eventually.

The proposed algorithm falls somewhere near the boundary between the robust algorithm approach and the correction approach aforementioned. It may be seen as a robust algorithm approach since the original algorithm has some natural tolerance to label noise, although it was not designed to handle this specific problem. In addition, it was also improved to dynamically discover and re-label label noise, thus stopping the noise propagation and allowing the algorithm to achieve higher classification accuracy. In this sense, this approach may be seem as belonging to the correction approach type. It is important to notice that this correction is a built-in feature. Both labeling unlabeled nodes and fixing label noise tasks run together in a single process.

Computer simulations presented in this paper show the effectiveness and robustness of the improved algorithm in the presence of high amounts of label noise. The classification accuracy achieved by the proposed method is compared with those achieved by all three previous versions and also with those achieved by some other representative graph-based semi-supervised learning methods \cite{Zhou2004,Zhu2002,Wang2008}. Both artificially generated and real-world data sets were used. Label noise was introduced in these data sets with increasing levels to discover how much label noise each algorithm can handle until the classification accuracy seriously drops.

This paper is organized as follows. An overview of the particle competition and cooperation approach is shown in Section \ref{sec:PCCOverview}. The proposed model is described in Section \ref{sec:ProposedModel}. In Section \ref{sec:ComputerSimulations}, we present computer simulations. Finally, in Section \ref{sec:Conclusions} we draw some conclusions.

\section{Particle Competition and Cooperation Overview}
\label{sec:PCCOverview}

In this section, we present an overview of the previous particle competition and cooperation models \cite{Breve2012TKDE,Breve2010IJCNN,Breve2012SBRN}. First, the vector-based data set is converted to a non-weighted and undirected graph. Each data instance becomes a graph node. Edges connecting the nodes are created according to the distance between the nodes in the data feature space. This graph generating process is described in Subsection \ref{sec:GraphConstruction}. Then, a particle is created for each labeled node. Particles with the same label belong to the same team and cooperate among themselves. On the other hand, particles with different labels compete against each other. When the system runs, the particles walk in the graph, selecting the next node to visit according to the rules described in Subsection \ref{sec:RandomGreedyWalk}. Each node has a set of domination levels, one level for each class of the problem. When a particle visits a node, it will increase its class domination level on that node, at the same time that it will decrease the domination level of the other classes. Each particle possesses a strength level, which lowers or raises according to the domination level of its class in the node it is being visited. Particles also have a distance table which they update dynamically as they walk on the graph. Nodes and particles dynamics are describe in Subsection \ref{sec:NodesParticlesDynamics}. The stop criterion is described in Subsection \ref{sec:StopCriterion}. At the end of the iterative process, each data item is labeled after the class with the highest domination level on it.

\subsection{Graph Construction}
\label{sec:GraphConstruction}

Consider a vector-based data set $\mathbf{\chi} = \{\mathbf{x}_1,\mathbf{x}_2,\dots,\mathbf{x}_l,\mathbf{x}_{l+1},\dots,\mathbf{x}_n\} \subset \mathds{R}^{m}$ with numerical attributes, and the corresponding label set $L = \{1,2,\dots,c\}$. The first $l$ points $x_i (i \leq l)$ are labeled as $y_i \in L$ and the remaining points $x_u (l < u \leq n)$ are unlabeled, i.e,  $y_u = \emptyset$. We define the graph $\mathbf{G} = (\mathbf{V},\mathbf{E})$. $\mathbf{V} = \{v_1,v_2,\dots,v_n\}$ is the set of nodes, where each one $v_i$ corresponds to a sample $\mathbf{x}_i \in \mathbf{\chi}$, and $\mathbf{E}$ is the set of edges $(v_i, v_j)$.

In \cite{Breve2012TKDE} and \cite{Breve2010IJCNN}, two nodes $v_i$ and $v_j$ are connected if the distance (usually the Euclidean distance) between $x_i$ and $x_j$ is below a given threshold $\sigma$. Since the threshold may be hard to define, another option is to connect $v_i$ and $v_j$ if $x_j$ is among the $k$-nearest neighbors of $x_i$ or vice-versa. Otherwise, $v_i$ and $v_j$ are disconnected. In \cite{Breve2012SBRN}, $v_i$ and $v_j$ are connected if $x_j$ is among the $k$-nearest neighbors of $x_i$ or vice-versa; or if $x_i$ and $x_j$ are both labeled instances with the same label. Otherwise, they are disconnected. This last rule was introduced to provide an easy and fast escape path to particles starting in nodes representing label noise samples. However, there is a side effect in this strategy, which will be discussed in Section \ref{sec:NewGraphConstruction}.

\subsection{Particles and Nodes Initialization}
\label{sec:ParticlesNodesInit}

For each labeled node $v_i \in \{v_1,v_2,\dots,v_l \}$ in the graph, which corresponds to a labeled data point $\mathbf{x}_i \in \{\mathbf{x}_1,\mathbf{x}_2,\dots,\mathbf{x}_l\}$, there is a particle $\rho_i \in  \{\rho_1,\rho_2,\dots,\rho_l\}$ which has $v_i$ as its initial position. $v_i$ is called the \emph{home node} of $\rho_i$.

Each particle $\rho_j$ holds two variables. The first one is $\rho_j^\omega(t) \in [0, 1]$ corresponding to the particle strength level, which indicates how much the particle is able to change the visited node levels at time $t+1$. The second variable is a distance table, i.e., a vector $\rho_j^\mathbf{d}(t) = \{\rho_j^{d_1}(t),\rho_j^{d_2}(t),\dots,\rho_j^{d_n}(t)\}$, where each element $\rho_j^{d_i}(t) \in [0, \quad n-1]$ corresponds to the distance dynamically measured between the particle's home node $v_j$ and the node $v_i$. Each particle $\rho_j$ is created with initial strength level set to maximum, $\rho_j^{\omega}(0)=1$. Particles begin their journey knowing only the distance to their corresponding labeled nodes, which is set to zero ($\rho_j^{d_i}=0$). Other distances are set to the largest possible value ($\rho_j^{d_i}=n-1$) if the graph is a single component.

Each node $v_i$ has a vector variable $v_i^\mathbf{\omega}(t) = \{v_i^{\omega_1}(t), v_i^{\omega_2}(t), \dots, v_i^{\omega_c}(t) \}$, where each element $v_i^{\omega_\ell}(t) \in [0, 1]$ corresponds to the domination level of team (class) $\ell$ over node $v_i$. For each node, the sum of the domination levels is always constant, $\sum_{\ell=1}^{c} v_i^{\omega_\ell}(t) = 1$. The initial domination levels are set differently for labeled nodes and unlabeled nodes. Labeled nodes begin fully dominated by the corresponding team (class). On the other hand, unlabeled nodes have all teams (classes) domination levels set equally. Therefore, for each node $v_i$, the initial levels of the domination vector $v_i^\mathbf{\omega}$ are set as follows:
\begin{equation}\label{eq:NodesInit}
    v_i^{\omega_\ell}(0) = \left\{
    \begin{array}{ccl}
        1 & & \mbox{if $y_i = \ell$} \\
        0 & & \mbox{if $y_i \neq \ell$ and $y_i \in L$} \\
        \frac{1}{c} & & \mbox{if $y_i = \emptyset$}
    \end{array}\right..
\end{equation}

\subsection{Random-Greedy Walk}
\label{sec:RandomGreedyWalk}

Particles walk in the graph trying to dominate as many nodes as possible, while preventing enemy particles from invading their territory. But how do they do that? This is the job of the two rules: \emph{random walk} and \emph{greedy walk}. They are used to determine which is the next node a particle will visit. In the \emph{random walk}, particles randomly chooses any neighbor to visit without concerning domination levels or distance from its home node. This rule is useful for exploration and acquisition of new nodes. Meanwhile, in the \emph{greedy walk}, particles prefer visiting nodes that have been already dominated by its own team and that are closer to their home nodes. This rule is useful for defense of its team's territory. Particles must exhibit both movements in order to achieve an equilibrium between exploratory and defensive behavior.

Therefore, in \emph{random walk} the particle $\rho_j$ moves to any node $v_i$ with the probabilities defined as:
\begin{equation}\label{eq:ProbRan}
    p(v_i|\rho_j) = \frac{W_{qi}}{\sum_{\mu=1}^{n}{W_{q \mu}}},
\end{equation}
where $q$ is the index of the current node of particle $\rho_j$, so $W_{qi}=1$ if there is an edge between the current node and any node $v_i$, and $W_{qi}=0$ otherwise. In \emph{greedy movement} the particle moves to a neighbor with probabilities defined according to its team domination level on that neighbor $\rho_j^{\omega_\ell}$ and inverse of the distance ($\rho_j^{d_i}$) from that neighbor $v_i$ to its home node $v_j$ by the following expression,
\begin{equation}\label{eq:ProbDet}
    p(v_i|\rho_j) = \frac{W_{qi} v_i^{\omega_\ell} \frac{1}{(1+\rho_j^{d_i})^2}} {\sum_{\mu=1}^{n}{W_{q\mu} v_\mu^{\omega_\ell}} \frac{1}{(1+\rho_j^{d_\mu})^2}}.
\end{equation}
where $q$ is the index of the current node of particle $\rho_j$ and $\ell = \rho_j^f$, where $\rho_j^f$ is the class label of particle $\rho_j$.

At each iteration, each particle has probability $p_{\textrm{grd}}$ to take greedy movement and probability $1-p_{\textrm{grd}}$ to take random movement, with $0 \leq p_{\textrm{grd}} \leq 1$. Once the random rule or greedy rule is determined, the neighbor node to be visited is chosen with probabilities defined by Eq. \eqref{eq:ProbRan} or Eq. \eqref{eq:ProbDet}, respectively.

When a particle visits a node, it updates the domination level on that node, its own strength and its distance table, as we will see later. But, after that, the particle only stays in the visited node until the next iteration if its team (class) domination level on that node is higher than those from all other teams (classes); otherwise, a shock happens and the particle is pushed back to its previous node until the next iteration.

In \cite{Breve2012SBRN}, the equations \eqref{eq:ProbRan} and \eqref{eq:ProbDet} were replaced by a single random-greedy equation:
\begin{equation}\label{eq:ProbMov}
    p(v_i|\rho_j) = 0.5 \left( \frac{W_{qi}}{\sum_{\mu=1}^{n}{W_{q \mu}}} +  \frac{W_{qi} v_i^{\omega_\ell} (1+\rho_j^{d_i})^{-2}} {\sum_{\mu=1}^{n}{W_{q\mu} v_i^{\omega_\ell}} (1+\rho_j^{d_i})^{-2}} \right),
\end{equation}
where $q$ is the index of the node currently being visited by particle $\rho_j$. This new random-greedy equation balances exploratory and defensive behavior.

\subsection{Nodes and Particles Dynamics}
\label{sec:NodesParticlesDynamics}

As mentioned before, at each iteration $t$, each particle $p_j$ chooses a neighbor node $v_i$ to visit. During this visit, particle $p_j$ updates the domination level of the neighbor node $v_i^{\omega_\ell}(t)$ as follows:
\begin{equation}\label{eq:UpdateNodesPot}
    v_i^{\omega_\ell}(t+1) = \left\{
    \begin{array}{l}
        \max\{0,v_i^{\omega_\ell}(t) - \frac{\Delta_{v} \rho_j^{\omega}(t)}{c-1}\} \\
        \quad \mbox{if $v_i$ is unlabeled and $\ell \neq \rho_j^f$} \\
        v_i^{\omega_\ell}(t) + \sum_{q \neq \ell}{v_i^{\omega_q}(t) - v_i^{\omega_q}(t+1)} \\
        \quad \mbox{if $v_i$ is unlabeled and $\ell = \rho_j^f$} \\
        v_i^{\omega_\ell}(t) \quad \mbox{if $v_i$ is labeled}
    \end{array}\right.,
\end{equation}
where $0 < \Delta_v \leq 1$ is a parameter to control changing rate of the domination levels and $\rho_j^f$ represents the class label of particle $\rho_j$. The update consists of particle $\rho_j$ changing the visited node $v_i$ by increasing the domination level of its team ($v_i^{\omega_\ell}$, $\ell=\rho_j^f$) while decreasing the domination levels of other teams ($v_i^{\omega_\ell}$, $\ell \neq \rho_j^f$)). In \cite{Breve2012TKDE}, there is an exception: the domination levels of labeled nodes are always fixed, assuming that their respective labels are always reliable.

When visiting a neighbor node, a particle will get weaker or stronger according to the domination level of its team in that node, after \eqref{eq:UpdateNodesPot} is applied. Therefore, at each iteration, a particle strength is updated:
\begin{equation}
    \label{eq:UpdatePartPot}
    \rho_j^{\omega}(t) = v_i^{\omega_\ell}(t),
\end{equation}
where $v_i$ is the node being visited. In \cite{Breve2010IJCNN} there is also a parameter $\Delta_{\rho}$ to control the amplitude of the particle strength change, so this Eq. \eqref{eq:UpdatePartPot} becomes
\begin{equation}
    \rho_j^{\omega}(t+1) = \rho_j^{\omega}(t) + \Delta_{\rho} (v_i^{\omega_\ell}(t+1) - \rho_j^{\omega}(t)).
\end{equation}

In \cite{Breve2010IJCNN}, there are also accumulated domination levels, which are defined as $v_i^\mathbf{\lambda}$, which is a vector $v_i^\mathbf{\lambda}(t) = \{v_i^{\lambda_1}(t),v_i^{\lambda_2}(t),\dots,v_i^{\lambda_c}(t)\}$ of the same size as $L$, and $v_i^{\lambda_\ell}(t) \in [0 \quad \infty]$ holds accumulated domination level by team $\ell$ over node $v_i$. At each iteration, for each selected node $v_i$ (in \emph{random movement}), the accumulated domination level $v_i^{\lambda_\ell}(t)$ is updated as follows::
\begin{equation}\label{eq:UpdateLongTermLevels}
    v_i^{\lambda_\ell}(t+1) = v_i^{\lambda_\ell}(t) + \rho_j^{\omega}(t)
\end{equation}
where $\ell$ is the class label of particle $\rho_j$.

During the visit the particle $\rho_j$ also updates its distance table $\rho_j^{d_k}(t)$ as follows:
\begin{equation}\label{eq:UpdatePartDist}
    \rho_j^{d_k}(t+1) = \left\{
    \begin{array}{cl}
        \rho_j^{d_i}(t) + 1 & \mbox{if } \rho_j^{d_i}(t) + 1 < \rho_j^{d_k}(t) \\
        \rho_j^{d_k}(t) & \mbox{otherwise}
    \end{array}\right.,
\end{equation}
where $\rho_j^{d_i}(t)$ and $\rho_j^{d_k}(t)$ are the distances to $\rho_j$ home node from the previous node and from the visited node, respectively.

Distance calculation is a dynamical process: particles have limited knowledge of the network, i.e., they do not know the connection pattern of nodes. Therefore, they assume all the nodes can be reached only with a number of steps as high as the total amount of nodes minus one ($n-1$) starting from its home node. Every time a particle chooses a neighbor node to visit, it will check the distance to that node in its distance table. If the distance on the table is higher than the distance it has from the previous node plus 1, it will update the table. In other words, unknown distances are calculated on the fly and updated as particles naturally find shorter paths while they walk. In \cite{Breve2012SBRN}, all particles from the same team share the same distance table.

\subsection{Stop Criterion}
\label{sec:StopCriterion}

In most scenarios, after a sufficient amount of iterations, most nodes will be locally dominated by a single team. Lets call this the equilibrium state. At this point, most nodes are unlikely to have major changes in their domination levels. However, some special nodes like the nodes on frontier regions, and nodes with label noise together with their closer neighbors are less stable and more susceptible to changes in the domination levels even after the equilibrium is reached. Therefore, we cannot expect full convergence of all node labels every time. Instead, we monitor the average maximum domination levels of the nodes ($\langle v_i^{\omega\ell} \rangle$, $\ell=\arg\max_q v_i^{\omega_q}$) and we keep track of the highest level they have achieved. This measure usually increases quickly at the beginning, then it slows down and oscillates around the maximum point. When there is no increase in this highest level achieved for a given amount of iterations ($\tau$) the algorithm is stopped. These $\tau$ iterations are needed because those special nodes usually require more iterations than others to provide more reliable labels. In other words, the particle competition and cooperation algorithms spend only a small portion of all the iterations to classify most nodes, and then they spend most of the remaining iterations ($\tau$) to classify the few remaining nodes. Using the algorithms proposed in \cite{Breve2012TKDE}, \cite{Breve2010IJCNN}, and \cite{Breve2012SBRN}, we usually set $\tau = \|\frac{\alpha n}{l}\|$, where $n$ is the network size, $l$ is the amount of labeled nodes (particles), and $\alpha$ is a constant. In this paper, we use $\alpha=2,000$. In our experiments values lower than that usually leads to lower classification accuracy, while values higher than that usually leads to no improvement in classification, but higher execution time.

When the algorithm stops completely, in \cite{Breve2012TKDE} and \cite{Breve2012SBRN}, each node is labeled or relabeled (only in \cite{Breve2012SBRN}) by the class which has the higher level of domination in it:
\begin{equation}
    \label{eq:LabelingNodes}
    y_i = \arg\max_{\ell} v_i^{\omega_\ell}.
\end{equation}

On the other hand, \cite{Breve2010IJCNN} uses the accumulated domination levels to label (or relabel) each node:
\begin{equation}
    \label{eq:LabelingNodesIJCNN2010}
    y_i = \arg\max_{\ell} v_i^{\lambda_\ell}.
\end{equation}

\section{The Proposed Model}
\label{sec:ProposedModel}

In this section, we present the features introduced in the particle competition and cooperation approach to minimize the effects of label noise. The new graph construction steps are described in Subsection \ref{sec:NewGraphConstruction}. The changes in particles and node initialization are describe in Subsection \ref{sec:NewParticlesNodesInit}.  The new particles and nodes dynamics are described in Subsection \ref{sec:NewNodesParticlesDynamics}. A rule to reset particles and nodes is discussed on Subsection \ref{sec:ResetRule}. Finally, an overview of the proposed algorithm is presented in Subsection \ref{sec:Algorithm}.

\subsection{Graph Construction}
\label{sec:NewGraphConstruction}

In Section \ref{sec:GraphConstruction} we described how the vector-based data set is converted to a non-weighted undirect graph in \cite{Breve2012TKDE}, \cite{Breve2010IJCNN}, and \cite{Breve2012SBRN}. Remember that Ref. \cite{Breve2012SBRN} introduced a rule to provide an easy and fast path to particles corresponding to nodes with class noise escape to their own neighborhood. However, there is a side effect in this strategy. The degrees of labeled nodes will increase according to the amount of labeled nodes of the same class, while the degrees of unlabeled nodes will depend only on the value of $k$ (from the $k$-nearest neighbors connections) and reciprocal connections. This fact may lead to graphs that labeled nodes have much higher degree than unlabeled ones. In this scenario, particles will spend too much time walking only on labeled nodes, which delays the algorithm stop and may also affect its classification accuracy. Therefore, in the proposed method we fix this problem by using a different strategy to connected label and unlabeled nodes. Here, each unlabeled node is connected to its $k$-nearest neighbors, no matter whether these neighbors are labeled or unlabeled (as in \cite{Breve2012TKDE}). On the other hand, labeled nodes are designed to prefer to connect to the $k$-nearest other labeled nodes from the same class. Only if $k-1$ is larger than the amount of nodes of the same class (lets call this amount $z$), the remaining connections will be made to the $(k-z)$-nearest neighbors no matter whether they are unlabeled nodes or labeled nodes but from other classes. Of course the connections are still reciprocal, as the network is undirected. Therefore labeled nodes will still be connected to their closest neighbors by reciprocity, but their degree will not be much larger than the degree of the unlabeled nodes.

\subsection{Particles and Nodes Initialization}
\label{sec:NewParticlesNodesInit}

Particles and nodes initialization is similar to the algorithm in \cite{Breve2012TKDE}, as described in Section \ref{sec:ParticlesNodesInit}. However, in the proposed method we introduce the overall domination levels $v_i^{\Omega}$, which have the same structure as $v_i^{\omega}$, but they are intended to keep overall information about nodes domination. These levels are all initially set to minimum:
\begin{equation}
\label{eq:OverallDominationLevelsInit}
    v_i^{\Omega_\ell} = 0,
\end{equation}
and they are updated when the reset rule is triggered, as it will be explained in Subsection \ref{sec:ResetRule}.

\subsection{Nodes and Particles Dynamics}
\label{sec:NewNodesParticlesDynamics}

In Section \ref{sec:NodesParticlesDynamics}, we mentioned that in \cite{Breve2012SBRN} all particles from the same team share the same distance table. This feature makes the particle's distance to any labeled node of the same class to be zero, which minimizes the importance of the home node and the particle's ``desire'' of going back to it, i.e., this rule allows particles to completely abandon their home nodes when they suffer from label noise, but their home nodes may be legitimate labeled nodes, which may lead to the territory switching phenomenon, where teams of particles may switch territory with another team, leading to major classification accuracy lost. To avoid this problem, we assume individual distance tables, i.e., each particle has its own table. Particles may still leave their home nodes faster than in \cite{Breve2012TKDE} and \cite{Breve2010IJCNN}, due to the changes proposed in the graph construction step, connecting labeled nodes even when they are not so close. But the new distance for each particle creates a stronger tie to the home node, minimizing territory switching phenomenon occurrences.

\subsection{Reset Rule}
\label{sec:ResetRule}

One of the problems observed in \cite{Breve2012SBRN} is the territory switching phenomenon, where a team of particles may switch territory with another team. This phenomenon leads to major classification accuracy lost, as two or more classes are usually almost entirely misclassified in these scenarios. In this paper, we made some enhancements to minimize this problem, including changes in graph construction and particles distance tables, which were described in Subsections \ref{sec:NewGraphConstruction} and \ref{sec:NewNodesParticlesDynamics}, respectively. Another novelty here is the introduction of a reset rule, which is used to reset all nodes domination levels and all particles position, strength, and distance tables from time to time. The reset is triggered when the highest level achieved by average maximum domination levels of the nodes ($\langle v_i^{\omega\ell} \rangle$, $\ell=\arg\max_q v_i^{\omega_q}$) has no increase in the last $\tau$ iterations. Here we change the definition of $\tau$ by introducing the new term $\beta$:
\begin{equation}
    \tau = \left\|\frac{\alpha n}{\beta l}\right\|
\end{equation}
where $n$ is the network size, $l$ is the amount of labeled nodes (particles), $\alpha$ is a constant and $\beta$ is the amount of resets that will be performed. Notice that as we increase $\beta$, $\tau$ decreases. Remember that, in Section \ref{sec:StopCriterion}, we explained that the amount of iterations to reach the equilibrium state (before the $\tau$ iterations take place) is usually much less than $\tau$. Therefore, the impact of the amount of resets ($\beta$) in execution time is negligible, and the execution time of the proposed method is nearly the same of the previous versions of the algorithm, given the same data set and parameters. In this paper, we set $\alpha=2,000$ (the same value used for the previous versions \cite{Breve2012TKDE,Breve2010IJCNN,Breve2012SBRN}). Increasing the value of $\beta$ minimizes the territory switching effect, increasing classification accuracy, but it also decreases the value of $\tau$, which may lead to lower classification accuracy in individual runs (between each reset). Therefore, $\beta$ must be carefully chosen. In our experiments, $\beta=10$ provided good shield against the territory switching effect without affecting classification accuracy of individual runs, so this value was used in all the experiments in this paper.

Each time the reset rule is triggered, all nodes' current domination levels are added to the nodes overall domination levels:
\begin{equation}
\label{eq:UpdateOverallDominationLevels}
    v_i^{\Omega} = v_i^{\Omega} + v_i^{\omega}(t).
\end{equation}
The overall domination levels are increased just before each reset. When the reset rule is triggered for the $\beta th$ time, the algorithm stops completely.

Thus, each node is labeled (or relabeled) by the class which has the higher overall level of domination in it:
\begin{equation}
    \label{eq:NewLabelingNodes}
    y_i = \arg\max_{\ell} v_i^{\Omega_\ell}
\end{equation}

\subsection{The Algorithm}
\label{sec:Algorithm}

Overall, the proposed algorithm can be outlined as described in Algorithm \ref{alg:Algorithm}.

\begin{algorithm}[h] \small
  Build the graph $G$ using the rules describe in Subsection \ref{sec:NewGraphConstruction}\;
  Set nodes' accumulated domination levels by using Eq. \eqref{eq:OverallDominationLevelsInit}\;
  \For{1 to $\beta$}
  {
    Set nodes' domination levels by using Eq. \eqref{eq:NodesInit}\;
    Set particles initial position, strength and distance tables by the rules described in Subsection \ref{sec:ParticlesNodesInit}\;
    \Repeat{the reset rule is triggered}
    {
        \For{each particle}
        {
            Select a neighbor node to visit by using  Eq. \eqref{eq:ProbMov}\;
            Update the visited node domination levels by using Eq. \eqref{eq:UpdateNodesPot}\;
            Update particle strength by using Eq. \eqref{eq:UpdatePartPot}\;
            Update particle distance tables by using Eq. \eqref{eq:UpdatePartDist}\;
        }
    }
    Update accumulated domination levels by using Eq. \eqref{eq:UpdateOverallDominationLevels}\;
  }
  Label each data item by using Eq. \eqref{eq:NewLabelingNodes}
  \caption{The Particle Competition and Cooperation Algorithm Enhanced to Minimize Effects of Label Noise}
  \label{alg:Algorithm}
\end{algorithm}

\section{Computer Simulations}
\label{sec:ComputerSimulations}

In this section, we present computer simulation results to show the effectiveness and robustness of the proposed method in the presence of label noise. We measure the classification accuracy of the proposed method when applied to artificial and real-world data sets, in which we introduce increasing amounts of label noise. The results of the proposed method, which we called the Label Noise Robust Particle Competition and Cooperation method (LNR-PCC), are compared to those achieved by three other representative graph-based semi-supervised learning methods: Local and Global Consistency (LGC) \cite{Zhou2004}, Label Propagation (LP) \cite{Zhu2002}, and Linear Neighborhood Propagation (LNP) \cite{Wang2008}. We also include the results achieved by three previous versions of the Particle Competition and Cooperation method: PCC-1 \cite{Breve2012TKDE}, PCC-2 \cite{Breve2010IJCNN}, and PCC-3 \cite{Breve2012SBRN}.

Regarding the parameters used in the algorithms in this experimental study, the following configuration is set. For the LGC and LNP methods, we have fixed $\alpha=0.99$, as done in \cite{Zhou2004} and \cite{Wang2008}, respectively. For PCC-1 and PCC-2 methods, we have fixed $p_{\textrm{grd}} = 0.5$, as done in \cite{Breve2010IJCNN}. Finally, $\Delta_v = 0.1$ is kept fixed in PCC-1, PCC-2, PCC-3, and LNR-PCC, as done in \cite{Breve2012TKDE,Breve2010IJCNN,Breve2012SBRN}. The most sensitive parameters, which are $\sigma$ of the LGC and the LP methods, and $k$ of LNP, PCC-1, PCC-2, PCC-3 and LNR-PCC methods, are all optimized using the genetic algorithm available in the Global Optimization Toolbox of MATLAB, aiming to minimize the classification error.

In all the simulations presented in this paper, we randomly select a subset of elements ($L \subset N$) to be presented to the algorithm with their labels (considered as labeled data instances), while the other elements in the data set are presented to the algorithm without labels (considered as unlabeled data instances). The only exception is the g241c data set, in which we use the labeled subsets shown in \cite{Chapelle2006}, instead of random selecting. In order to test robustness to label noise, we randomly choose $q$ elements from the labeled subset $L$ ($Q \subset L$) to have their labels changed to any of the other classes chosen randomly for each sample, thus producing label noise. These label noise subsets are generated with increasing sizes, $q/l = \{0.00, 0.05, 0.10, \dots\}$, until the labeled subset is no better than a random labeled subset. For instance, in a four classes problem with equiprobable classes, one can expect $\sim 25\%$ classification accuracy if the samples are labeled randomly. Therefore, there is no point in using a labeled subset in this scenario if the label noise amount is higher than $75\%$.

\begin{figure}
\centering
\includegraphics[width=12cm]{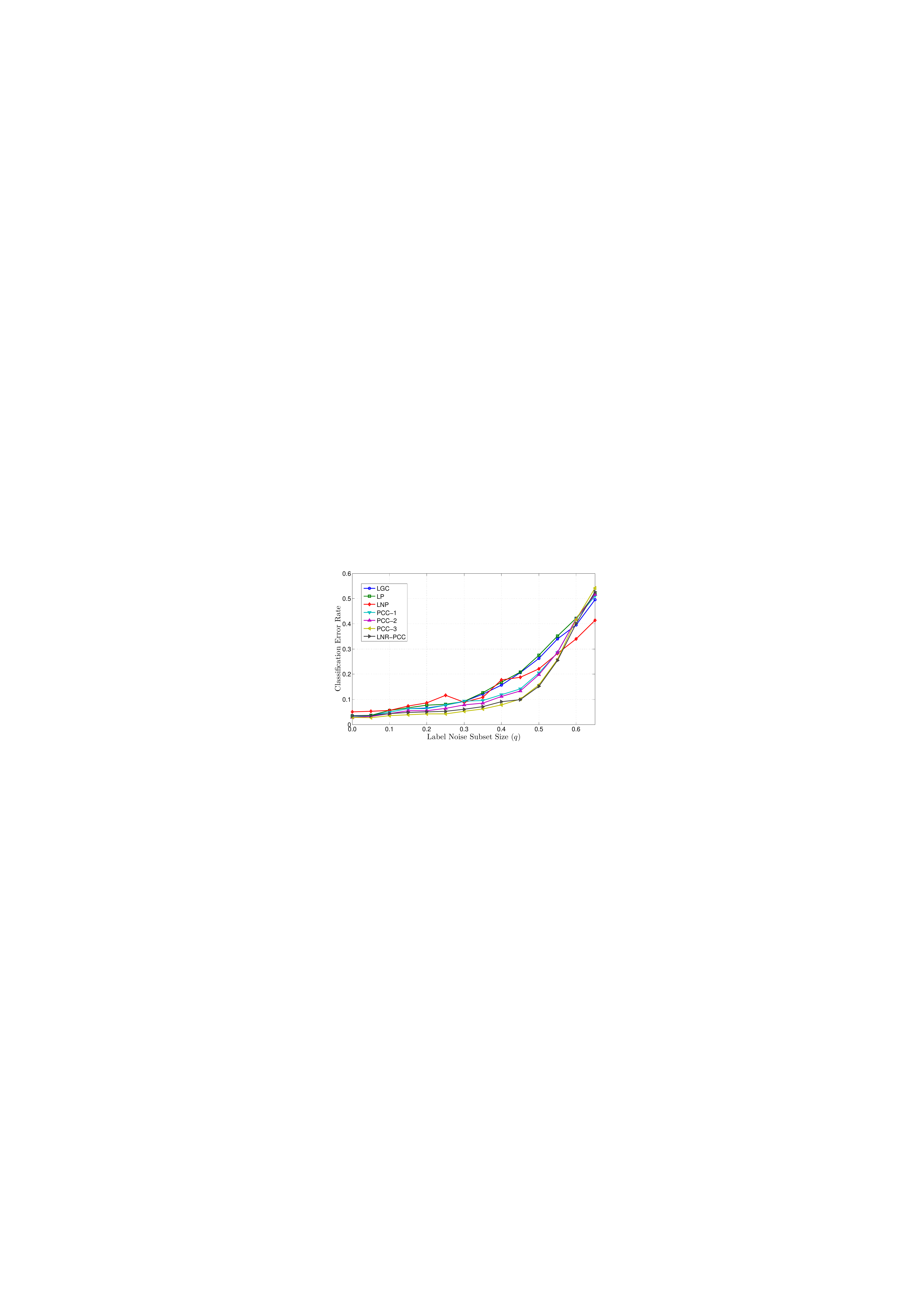}
\caption{Classification error rate in the Iris data set \cite{UCI2013} with different label noise subset ($Q$) sizes. Each data point is the average of $50$ executions with different $L$ and $Q$ subsets.}
\label{fig:iris}
\end{figure}

Figure \ref{fig:iris} and Table \ref{tab:iris} show the classification accuracy comparison when the semi-supervised learning graph-based methods are applied to the Iris Data Set \cite{UCI2013}, which has $150$ elements distributed in $3$ classes. $40$ data items are randomly chosen to compose the labeled subset. When the label noise subset size is small ($\leq 40\%$), the proposed method is outperformed only by PCC-3. When label noise reaches critical levels ($45\% \sim 55\%$), the proposed algorithm performs better than all the others.

\begin{table}
  \centering
  \caption{Classification error rate in the Iris data set \cite{UCI2013} with different label noise subset ($Q$) sizes. Each value is the average of $50$ executions with different $L$ and $Q$ subsets.}
    \begin{tabular}{cccccccc}
    \toprule
    \emph{Q} size & LGC   & LP    & LNP   & PCC-1 & PCC-2 & PCC-3 & LNR-PCC \\
    \midrule
    0.00  & 0.0353 & 0.0340 & 0.0498 & 0.0290 & 0.0286 & 0.0279 & 0.0345 \\
    0.50  & 0.0360 & 0.0364 & 0.0522 & 0.0346 & 0.0314 & 0.0273 & 0.0355 \\
    0.10  & 0.0495 & 0.0562 & 0.0555 & 0.0488 & 0.0431 & 0.0357 & 0.0427 \\
    0.15  & 0.0627 & 0.0667 & 0.0738 & 0.0629 & 0.0551 & 0.0386 & 0.0481 \\
    0.20  & 0.0655 & 0.0762 & 0.0862 & 0.0634 & 0.0543 & 0.0420 & 0.0499 \\
    0.25  & 0.0769 & 0.0807 & 0.1156 & 0.0770 & 0.0641 & 0.0427 & 0.0523 \\
    0.30  & 0.0922 & 0.0920 & 0.0887 & 0.0926 & 0.0784 & 0.0520 & 0.0603 \\
    0.35  & 0.1202 & 0.1264 & 0.1080 & 0.0954 & 0.0845 & 0.0618 & 0.0705 \\
    0.40  & 0.1565 & 0.1675 & 0.1769 & 0.1184 & 0.1114 & 0.0775 & 0.0902 \\
    0.45  & 0.2056 & 0.2078 & 0.1880 & 0.1415 & 0.1333 & 0.1004 & 0.0989 \\
    0.50  & 0.2629 & 0.2749 & 0.2211 & 0.2042 & 0.1982 & 0.1575 & 0.1512 \\
    0.55  & 0.3393 & 0.3515 & 0.2824 & 0.2851 & 0.2862 & 0.2593 & 0.2549 \\
    0.60  & 0.3953 & 0.4213 & 0.3400 & 0.4180 & 0.4198 & 0.4169 & 0.4008 \\
    0.65  & 0.4955 & 0.5171 & 0.4133 & 0.5110 & 0.5162 & 0.5419 & 0.5271 \\
    \bottomrule
    \end{tabular}%
  \label{tab:iris}%
\end{table}%

\begin{figure}
\centering
\includegraphics[width=12cm]{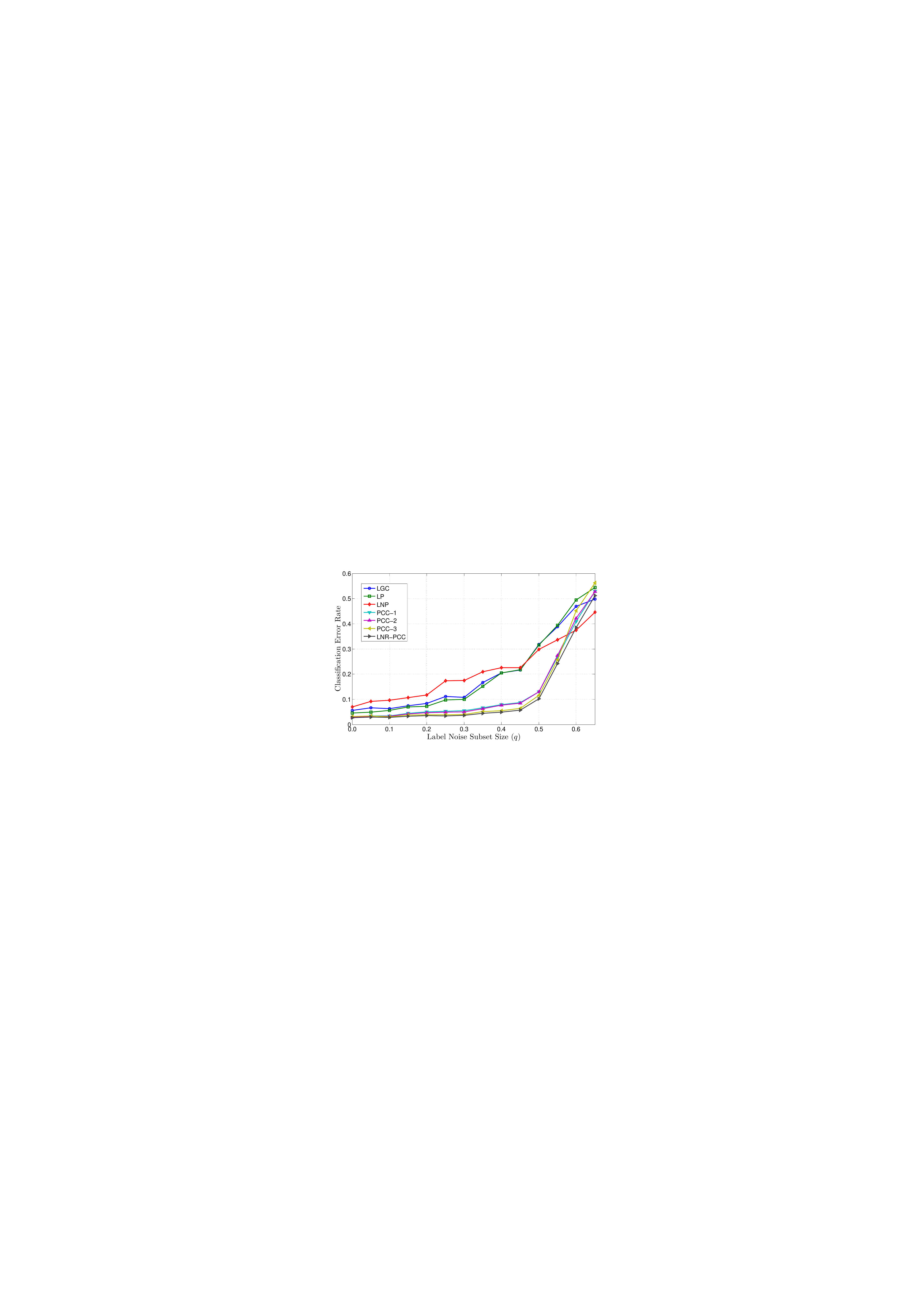}
\caption{Classification error rate in the Wine data set \cite{UCI2013} with different label noise subset ($Q$) sizes. Each data point is the average of $50$ executions with different $L$ and $Q$ subsets.}
\label{fig:wine}
\end{figure}

Figure \ref{fig:wine} and Table \ref{tab:wine} show the classification accuracy comparison when the methods are applied to the Wine Data Set \cite{UCI2013}, which has $178$ samples distributed in $3$ classes. $40$ samples are randomly chosen to compose the labeled subset. When all the samples in the labeled subset are correctly labeled, the proposed algorithm already performs better than all the others. As the mislabeled subset increases, this difference becomes higher because the classification error rates of the other algorithms increase more quickly than the the proposed method. Interestingly, the proposed algorithm seems not to be affected by up to $30\%$ of label noise. The classification accuracy begins to drop from $35\%$ label noise and beyond. However, this is the region where the proposed method increases its advantage over the others. When $50\%$ of the labeled subset is affected by label noise, the proposed method impressively made only around one third of the amount of classification mistakes made by LGC, LP, and LNP methods.

\begin{table}
  \centering
  \caption{Classification error rate in the Wine data set \cite{UCI2013} with different label noise subset ($Q$) sizes. Each data point is the average of $50$ executions with different $L$ and $Q$ subsets.}
    \begin{tabular}{cccccccc}
    \toprule
    \emph{Q} size & LGC   & LP    & LNP   & PCC-1 & PCC-2 & PCC-3 & LNR-PCC \\
    \midrule
    0.00  & 0.0562 & 0.0459 & 0.0699 & 0.0288 & 0.0289 & 0.0317 & 0.0274 \\
    0.05  & 0.0659 & 0.0490 & 0.0919 & 0.0346 & 0.0331 & 0.0343 & 0.0296 \\
    0.10  & 0.0630 & 0.0555 & 0.0968 & 0.0346 & 0.0326 & 0.0308 & 0.0284 \\
    0.15  & 0.0739 & 0.0697 & 0.1067 & 0.0444 & 0.0426 & 0.0372 & 0.0329 \\
    0.20  & 0.0838 & 0.0717 & 0.1168 & 0.0502 & 0.0464 & 0.0385 & 0.0351 \\
    0.25  & 0.1114 & 0.0975 & 0.1738 & 0.0525 & 0.0490 & 0.0383 & 0.0345 \\
    0.30  & 0.1084 & 0.1001 & 0.1754 & 0.0545 & 0.0506 & 0.0404 & 0.0360 \\
    0.35  & 0.1672 & 0.1522 & 0.2091 & 0.0665 & 0.0629 & 0.0518 & 0.0448 \\
    0.40  & 0.2046 & 0.2054 & 0.2257 & 0.0789 & 0.0763 & 0.0546 & 0.0495 \\
    0.45  & 0.2167 & 0.2171 & 0.2251 & 0.0876 & 0.0845 & 0.0641 & 0.0567 \\
    0.50  & 0.3178 & 0.3152 & 0.2983 & 0.1300 & 0.1301 & 0.1156 & 0.1017 \\
    0.55  & 0.3883 & 0.3942 & 0.3362 & 0.2682 & 0.2735 & 0.2568 & 0.2415 \\
    0.60  & 0.4690 & 0.4945 & 0.3745 & 0.4091 & 0.4215 & 0.4512 & 0.3843 \\
    0.65  & 0.4984 & 0.5445 & 0.4458 & 0.5294 & 0.5275 & 0.5628 & 0.5117 \\
    \bottomrule
    \end{tabular}%
  \label{tab:wine}%
\end{table}%

\begin{figure}
\centering
\includegraphics[width=12cm]{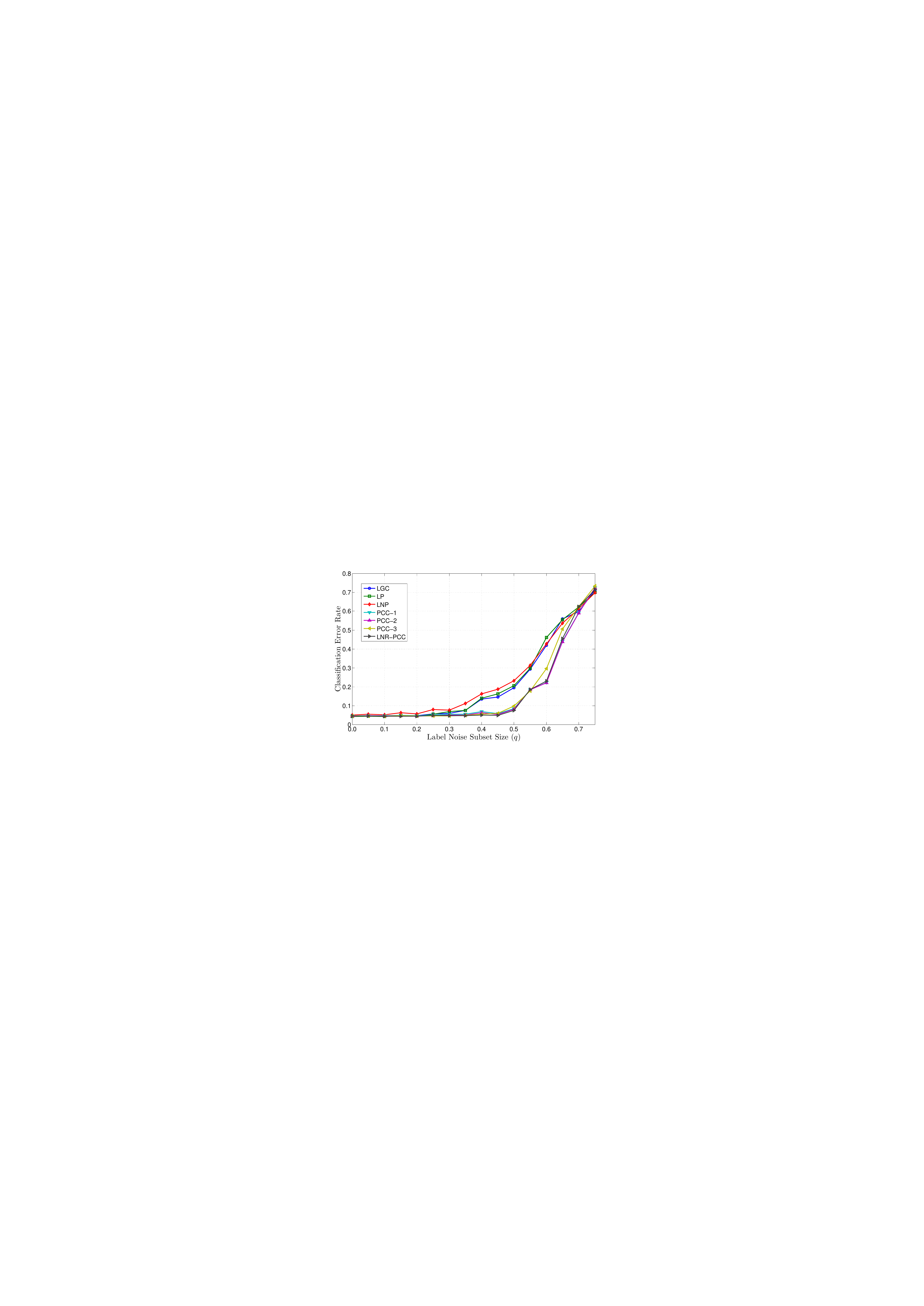}
\caption{Classification error rate in the data set with 4 normally distributed classes with different label noise subset sizes. Each data point is the average of $20$ executions with different $L$ and $Q$ subsets.}
\label{fig:4gauss}
\end{figure}

Figure \ref{fig:4gauss} and Table \ref{tab:4gauss} show the classification accuracy comparison when the semi-supervised learning graph-based methods are applied to an artificial data set with $1,000$ elements equally divided into $4$ normally distributed classes (Gaussian distribution). This data set was generated with function \texttt{gauss} from PRTools \cite{PRTools2007}. $50$ samples are randomly chosen to build the labeled subset. When there is no label noise, all the methods have similar classification accuracy, except LNP. As the label noise subset increases, the classification error rates begin to slowly raise. The PCC methods show their advantage by keeping nearly the same classification accuracy from $0\%$ to $45\%$ label noise, while the other methods drop their classification accuracy earlier ($30\% \sim 35\%$). In the range from $35\%$ to $50\%$ label noise, the proposed method outperformed all the others.

\begin{table}
  \centering
  \caption{Classification error rate in the data set with 4 normally distributed classes with different label noise subset sizes. Each data point is the average of $20$ executions with different $L$ and $Q$ subsets.}
    \begin{tabular}{cccccccc}
    \toprule
    \emph{Q} size & LGC   & LP    & LNP   & PCC-1 & PCC-2 & PCC-3 & LNR-PCC \\
    \midrule
    0.00  & 0.0457 & 0.0419 & 0.0503 & 0.0447 & 0.0442 & 0.0460 & 0.0453 \\
    0.05  & 0.0465 & 0.0435 & 0.0539 & 0.0471 & 0.0462 & 0.0457 & 0.0460 \\
    0.10  & 0.0444 & 0.0427 & 0.0511 & 0.0466 & 0.0454 & 0.0438 & 0.0439 \\
    0.15  & 0.0492 & 0.0484 & 0.0629 & 0.0490 & 0.0467 & 0.0462 & 0.0445 \\
    0.20  & 0.0463 & 0.0444 & 0.0559 & 0.0475 & 0.0448 & 0.0449 & 0.0433 \\
    0.25  & 0.0557 & 0.0544 & 0.0799 & 0.0505 & 0.0473 & 0.0450 & 0.0479 \\
    0.30  & 0.0597 & 0.0683 & 0.0768 & 0.0541 & 0.0494 & 0.0452 & 0.0462 \\
    0.35  & 0.0751 & 0.0751 & 0.1109 & 0.0542 & 0.0495 & 0.0483 & 0.0464 \\
    0.40  & 0.1346 & 0.1387 & 0.1631 & 0.0692 & 0.0627 & 0.0544 & 0.0499 \\
    0.45  & 0.1446 & 0.1622 & 0.1875 & 0.0575 & 0.0538 & 0.0591 & 0.0480 \\
    0.50  & 0.1942 & 0.2061 & 0.2311 & 0.0844 & 0.0813 & 0.0982 & 0.0745 \\
    0.55  & 0.2936 & 0.2978 & 0.3133 & 0.1840 & 0.1833 & 0.1781 & 0.1853 \\
    0.60  & 0.4193 & 0.4603 & 0.4263 & 0.2224 & 0.2215 & 0.2958 & 0.2296 \\
    0.65  & 0.5586 & 0.5551 & 0.5365 & 0.4406 & 0.4377 & 0.5051 & 0.4548 \\
    0.70  & 0.5973 & 0.6229 & 0.6126 & 0.5912 & 0.5905 & 0.6236 & 0.6240 \\
    0.75  & 0.7112 & 0.6999 & 0.6981 & 0.7241 & 0.7233 & 0.7330 & 0.7147 \\
    \bottomrule
    \end{tabular}%
  \label{tab:4gauss}%
\end{table}%

\begin{figure}
\centering
\includegraphics[width=12cm]{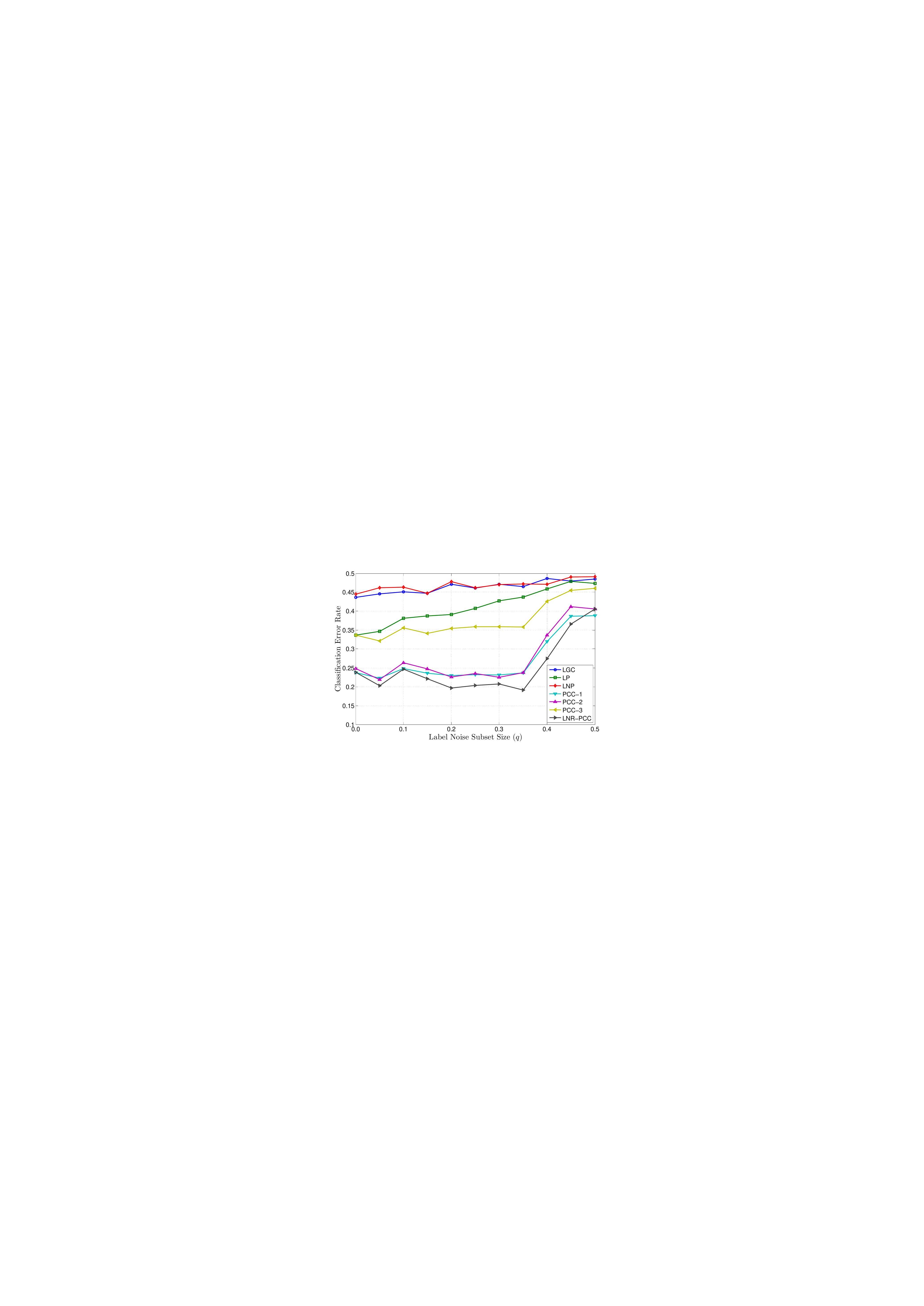}
\caption{Classification error rate in the g241c data set \cite{Chapelle2006} with different label noise subset ($Q$) sizes. Each data point is the average of $12$ executions with different $L$ and $Q$ subsets.}
\label{fig:g241c}
\end{figure}

Figure \ref{fig:g241c} and Table \ref{tab:g241c} show the classification accuracy comparison when the learning methods are applied to the g241c data set \cite{Chapelle2006}. The g241c data set is composed by $1500$ samples divided into $2$ classes. There are $12$ different labeled subsets, each of them containing $100$ samples, as provided in \cite{Chapelle2006}. From Figure \ref{fig:g241c} analysis, we see that the proposed method is better than all the others in the presence of $5\%$ to $45\%$ label noise.

\begin{table}
  \centering
  \caption{Classification error rate in the g241c data set \cite{Chapelle2006} with different label noise subset ($Q$) sizes. Each data point is the average of $12$ executions with different $L$ and $Q$ subsets.}
    \begin{tabular}{cccccccc}
    \toprule
    \emph{Q} size & LGC   & LP    & LNP   & PCC-1 & PCC-2 & PCC-3 & LNR-PCC \\
    \midrule
    0.00  & 0.4364 & 0.3362 & 0.4451 & 0.2372 & 0.2478 & 0.3362 & 0.2378 \\
    0.05  & 0.4455 & 0.3468 & 0.4615 & 0.2230 & 0.2187 & 0.3208 & 0.2028 \\
    0.10  & 0.4510 & 0.3810 & 0.4636 & 0.2482 & 0.2637 & 0.3560 & 0.2465 \\
    0.15  & 0.4470 & 0.3870 & 0.4476 & 0.2357 & 0.2471 & 0.3413 & 0.2213 \\
    0.20  & 0.4714 & 0.3913 & 0.4782 & 0.2299 & 0.2257 & 0.3540 & 0.1968 \\
    0.25  & 0.4612 & 0.4075 & 0.4621 & 0.2320 & 0.2350 & 0.3585 & 0.2031 \\
    0.30  & 0.4714 & 0.4276 & 0.4701 & 0.2313 & 0.2247 & 0.3586 & 0.2077 \\
    0.35  & 0.4649 & 0.4371 & 0.4716 & 0.2369 & 0.2372 & 0.3583 & 0.1913 \\
    0.40  & 0.4864 & 0.4586 & 0.4708 & 0.3196 & 0.3368 & 0.4260 & 0.2743 \\
    0.45  & 0.4800 & 0.4786 & 0.4901 & 0.3868 & 0.4122 & 0.4547 & 0.3656 \\
    0.50  & 0.4848 & 0.4734 & 0.4911 & 0.3878 & 0.4060 & 0.4603 & 0.4057 \\
    \bottomrule
    \end{tabular}%
  \label{tab:g241c}%
\end{table}%

\begin{figure}
\centering
\includegraphics[width=12cm]{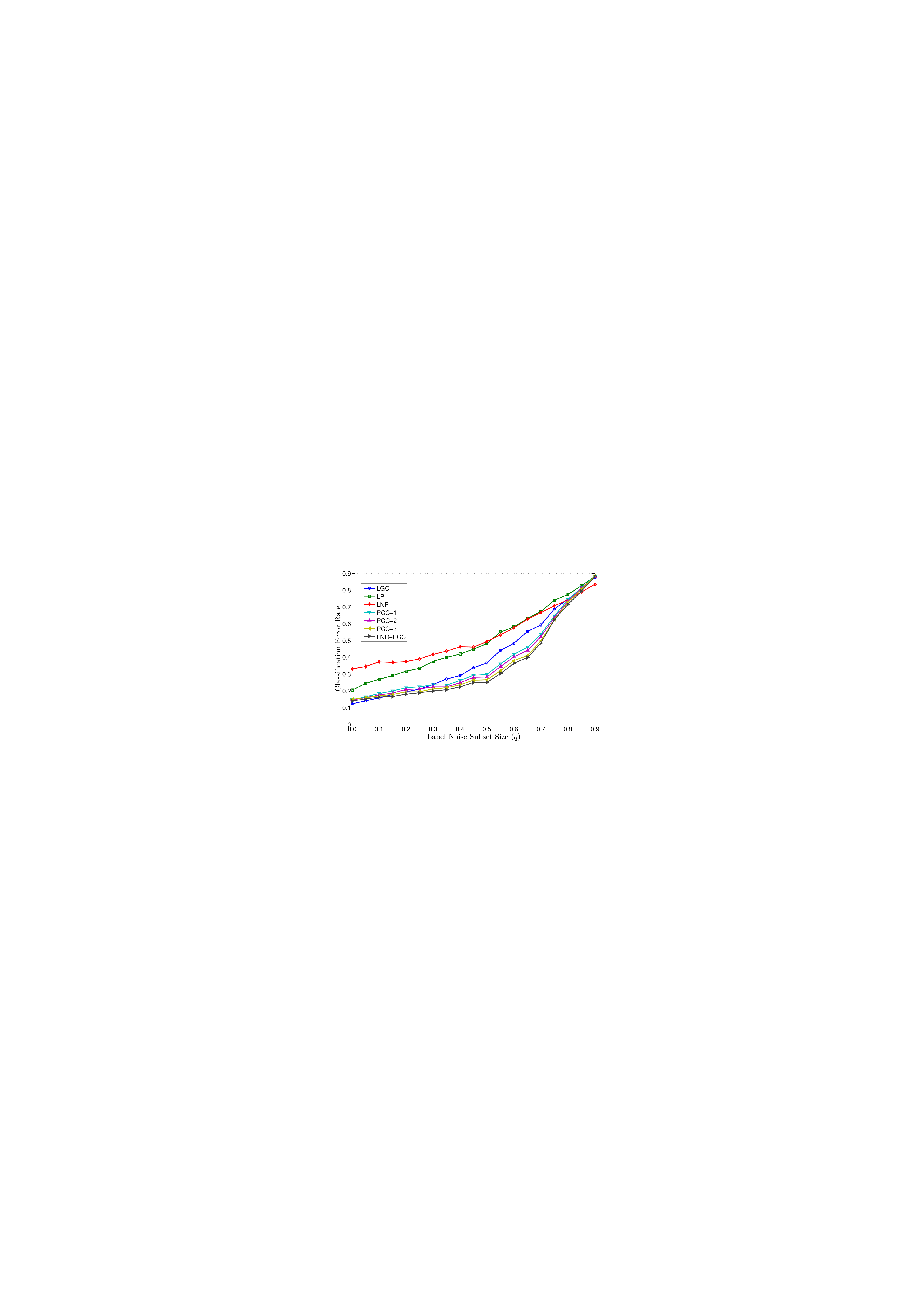}
\caption{Classification error rate in the Semeion Handwritten Digit data set  \cite{Semeion1,Semeion2} with different label noise subset ($Q$) sizes. Each data point is the average of $20$ executions with different $L$ and $Q$ subsets.}
\label{fig:semeion}
\end{figure}

Figure \ref{fig:semeion} and Table \ref{tab:semeion} show the classification accuracy comparison when the learning methods are applied to the Semeion Handwritten Digit data set \cite{Semeion1,Semeion2}, which has $1,593$ samples distributed in $10$ classes. $159$ samples are randomly chosen to compose the labeled subset. Only LGC is better than the proposed method when label noise affects $10\%$ or less of the labeled data. However, when $15\%$ to $80\%$ of the labeled subset is affected by label noise, the proposed method is better than all the others.

\begin{table}
  \centering
  \caption{Classification error rate in the Semeion Handwritten Digit data set  \cite{Semeion1,Semeion2} with different label noise subset ($Q$) sizes. Each data point is the average of $20$ executions with different $L$ and $Q$ subsets.}
    \begin{tabular}{cccccccc}
    \toprule
    \emph{Q} size & LGC   & LP    & LNP   & PCC-1 & PCC-2 & PCC-3 & LNR-PCC \\
    \midrule
    0.00  & 0.1238 & 0.2050 & 0.3307 & 0.1475 & 0.1459 & 0.1492 & 0.1414 \\
    0.05  & 0.1408 & 0.2443 & 0.3449 & 0.1658 & 0.1594 & 0.1596 & 0.1511 \\
    0.10  & 0.1582 & 0.2683 & 0.3722 & 0.1836 & 0.1763 & 0.1704 & 0.1628 \\
    0.15  & 0.1788 & 0.2912 & 0.3692 & 0.1994 & 0.1883 & 0.1786 & 0.1670 \\
    0.20  & 0.1945 & 0.3176 & 0.3747 & 0.2197 & 0.2080 & 0.1951 & 0.1809 \\
    0.25  & 0.2099 & 0.3354 & 0.3897 & 0.2236 & 0.2119 & 0.1944 & 0.1900 \\
    0.30  & 0.2381 & 0.3763 & 0.4185 & 0.2365 & 0.2245 & 0.2116 & 0.2003 \\
    0.35  & 0.2712 & 0.3996 & 0.4377 & 0.2343 & 0.2243 & 0.2183 & 0.2062 \\
    0.40  & 0.2910 & 0.4189 & 0.4623 & 0.2629 & 0.2486 & 0.2359 & 0.2241 \\
    0.45  & 0.3385 & 0.4488 & 0.4605 & 0.2930 & 0.2813 & 0.2641 & 0.2509 \\
    0.50  & 0.3654 & 0.4812 & 0.4946 & 0.2992 & 0.2831 & 0.2650 & 0.2502 \\
    0.55  & 0.4424 & 0.5509 & 0.5344 & 0.3610 & 0.3460 & 0.3211 & 0.3032 \\
    0.60  & 0.4846 & 0.5808 & 0.5747 & 0.4178 & 0.4028 & 0.3803 & 0.3639 \\
    0.65  & 0.5553 & 0.6328 & 0.6275 & 0.4609 & 0.4425 & 0.4108 & 0.3982 \\
    0.70  & 0.5934 & 0.6721 & 0.6661 & 0.5375 & 0.5240 & 0.4977 & 0.4849 \\
    0.75  & 0.6888 & 0.7393 & 0.7065 & 0.6472 & 0.6356 & 0.6258 & 0.6239 \\
    0.80  & 0.7473 & 0.7749 & 0.7415 & 0.7434 & 0.7343 & 0.7282 & 0.7151 \\
    0.85  & 0.8062 & 0.8262 & 0.7892 & 0.8144 & 0.8109 & 0.8053 & 0.7931 \\
    0.90  & 0.8748 & 0.8813 & 0.8343 & 0.8836 & 0.8803 & 0.8857 & 0.8810 \\
    \bottomrule
    \end{tabular}%
  \label{tab:semeion}%
\end{table}%

\begin{figure}
\centering
\includegraphics[width=12cm]{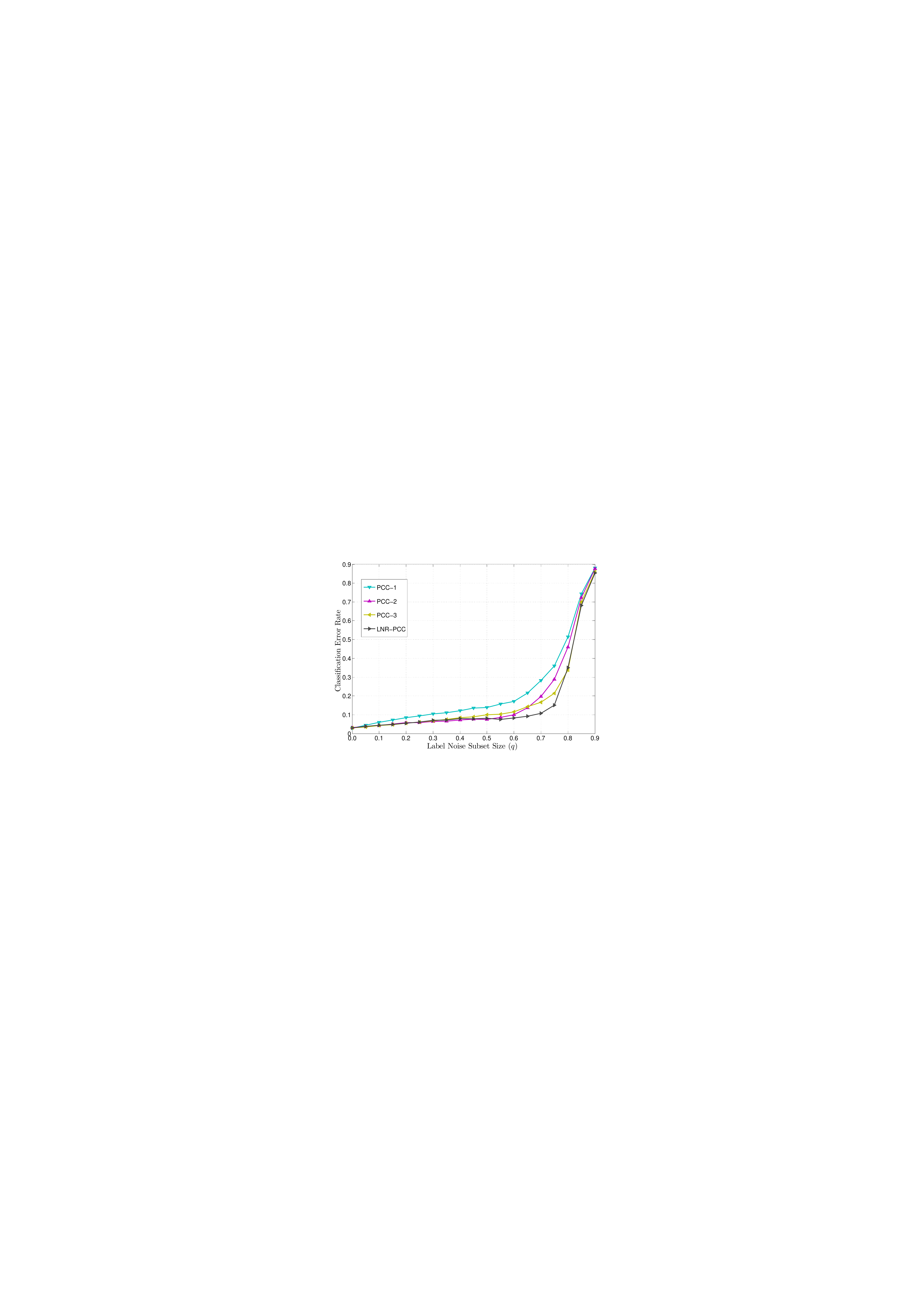}
\caption{Classification error rate in the Optical Recognition of Handwritten Digits data set \cite{UCI2013} with different label noise subset ($Q$) sizes. Each data point is the average of $5$ executions with different $L$ and $Q$ subsets.}
\label{fig:optdigits}
\end{figure}

Finally, Figure \ref{fig:optdigits} and Table \ref{tab:optdigits} show the classification accuracy comparison when the learning methods are applied to the Optical Recognition of Handwritten Digits data set \cite{UCI2013}, which has $5,620$ samples distributed in $10$ classes. $562$ samples are randomly chosen to compose the labeled subset. LGC, LP, and LNP methods were not applied to this data set due to the prohibitive execution time they would take. The proposed method is better than all previous versions in the presence of $55\%$ to $75\%$ label noise.

\begin{table}
  \centering
  \caption{Classification error rate in the Optical Recognition of Handwritten Digits data set \cite{UCI2013} with different label noise subset ($Q$) sizes. Each data point is the average of $5$ executions with different $L$ and $Q$ subsets.}
    \begin{tabular}{ccccc}
    \toprule
    \emph{Q} size & PCC-1 & PCC-2 & PCC-3 & LNR-PCC \\
    \midrule
    0.00  & 0.0292 & 0.0320 & 0.0295 & 0.0308 \\
    0.05  & 0.0445 & 0.0378 & 0.0349 & 0.0384 \\
    0.10  & 0.0591 & 0.0427 & 0.0419 & 0.0438 \\
    0.15  & 0.0725 & 0.0505 & 0.0478 & 0.0490 \\
    0.20  & 0.0838 & 0.0577 & 0.0544 & 0.0548 \\
    0.25  & 0.0931 & 0.0585 & 0.0611 & 0.0616 \\
    0.30  & 0.1038 & 0.0646 & 0.0672 & 0.0706 \\
    0.35  & 0.1105 & 0.0663 & 0.0753 & 0.0722 \\
    0.40  & 0.1211 & 0.0722 & 0.0840 & 0.0802 \\
    0.45  & 0.1347 & 0.0770 & 0.0883 & 0.0775 \\
    0.50  & 0.1379 & 0.0744 & 0.0999 & 0.0805 \\
    0.55  & 0.1573 & 0.0858 & 0.1029 & 0.0750 \\
    0.60  & 0.1716 & 0.0992 & 0.1157 & 0.0825 \\
    0.65  & 0.2158 & 0.1369 & 0.1425 & 0.0922 \\
    0.70  & 0.2823 & 0.1979 & 0.1662 & 0.1074 \\
    0.75  & 0.3598 & 0.2878 & 0.2135 & 0.1512 \\
    0.80  & 0.5136 & 0.4603 & 0.3383 & 0.3509 \\
    0.85  & 0.7434 & 0.7227 & 0.7000 & 0.6808 \\
    0.90  & 0.8810 & 0.8775 & 0.8604 & 0.8541 \\
    \bottomrule
    \end{tabular}%
  \label{tab:optdigits}%
\end{table}%

Each data point (each size of the label noise subset $Q$) in the Figures \ref{fig:iris} to \ref{fig:optdigits} curves is the average value from $5$ to $50$ executions (depending on the data set) with different elements in both labeled subset $L$ and label noise subset $Q$. Notice that the parameter optimization procedure is executed for each of these executions. Thus, each average value obtained by LGC, LP, and LNP is actually the average of the $5$ to $50$ best values obtained from the corresponding optimization process. On the other hand, for the PCC methods, the best value in each optimization process is discarded. Instead, the optimized parameters are used in another $20$ executions, since those are stochastic algorithms and the best classification accuracy obtained during the optimization process might be too optimistic. Therefore, for the PCC methods, each data point in the figures curves is actually the average value of $100$ to $1000$ executions ($5$ to $50$ configurations of elements in subsets $L$ and $Q$, and $20$ repetitions on each specific configuration).

\begin{figure}
\centering
\includegraphics[width=12cm]{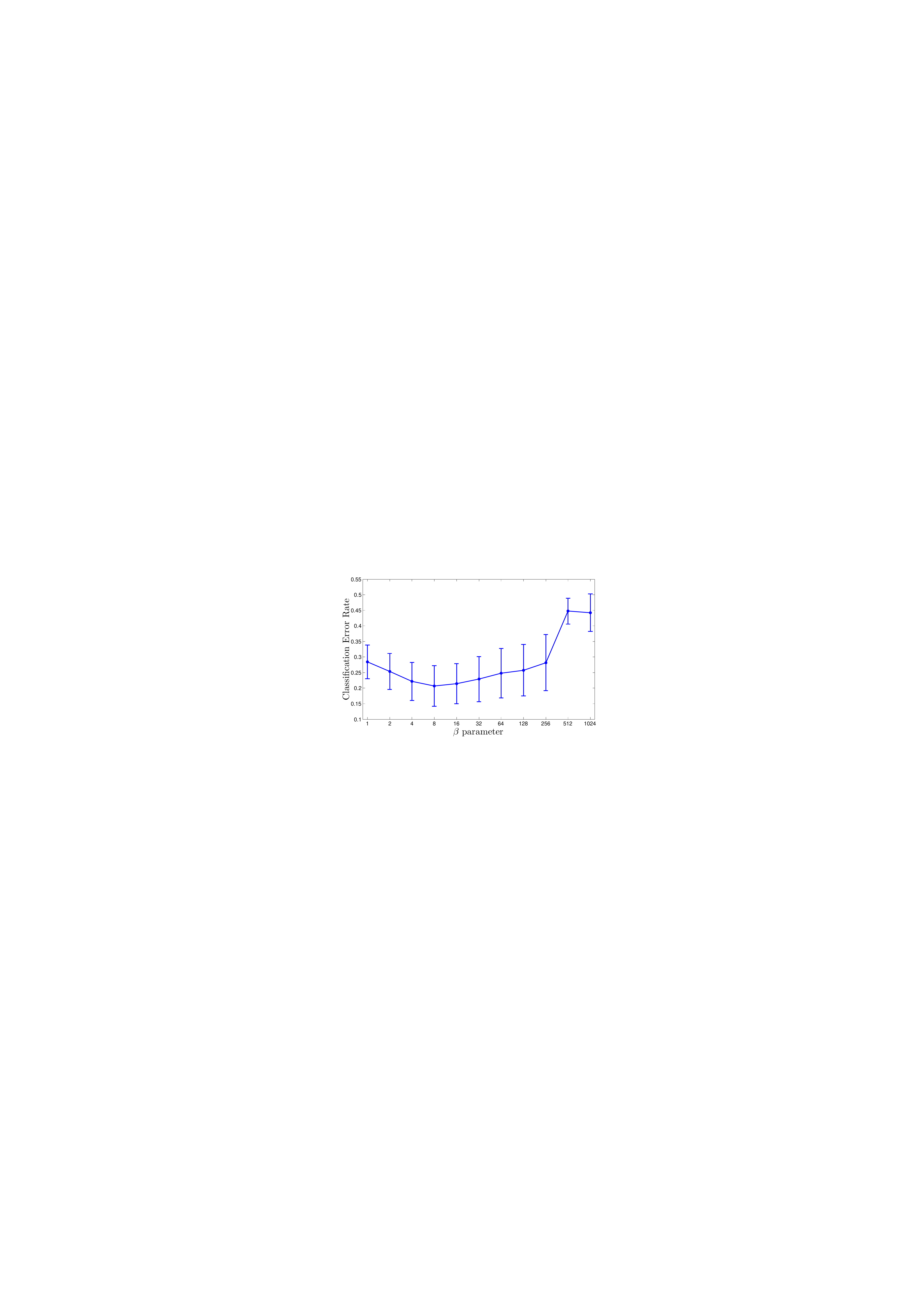}
\caption{Classification error rate and standard deviation with LNR-PCC using different values for $\beta$ parameter, applied to the g241c data set \cite{Chapelle2006} with $100$ labeled nodes, from which $35$ are incorrectly labeled. Each data point is the average of $240$ executions, $20$ on each of the $12$ labeled subsets defined by \cite{Chapelle2006}.}
\label{fig:beta-g241c}
\end{figure}

\begin{figure}
\centering
\includegraphics[width=12cm]{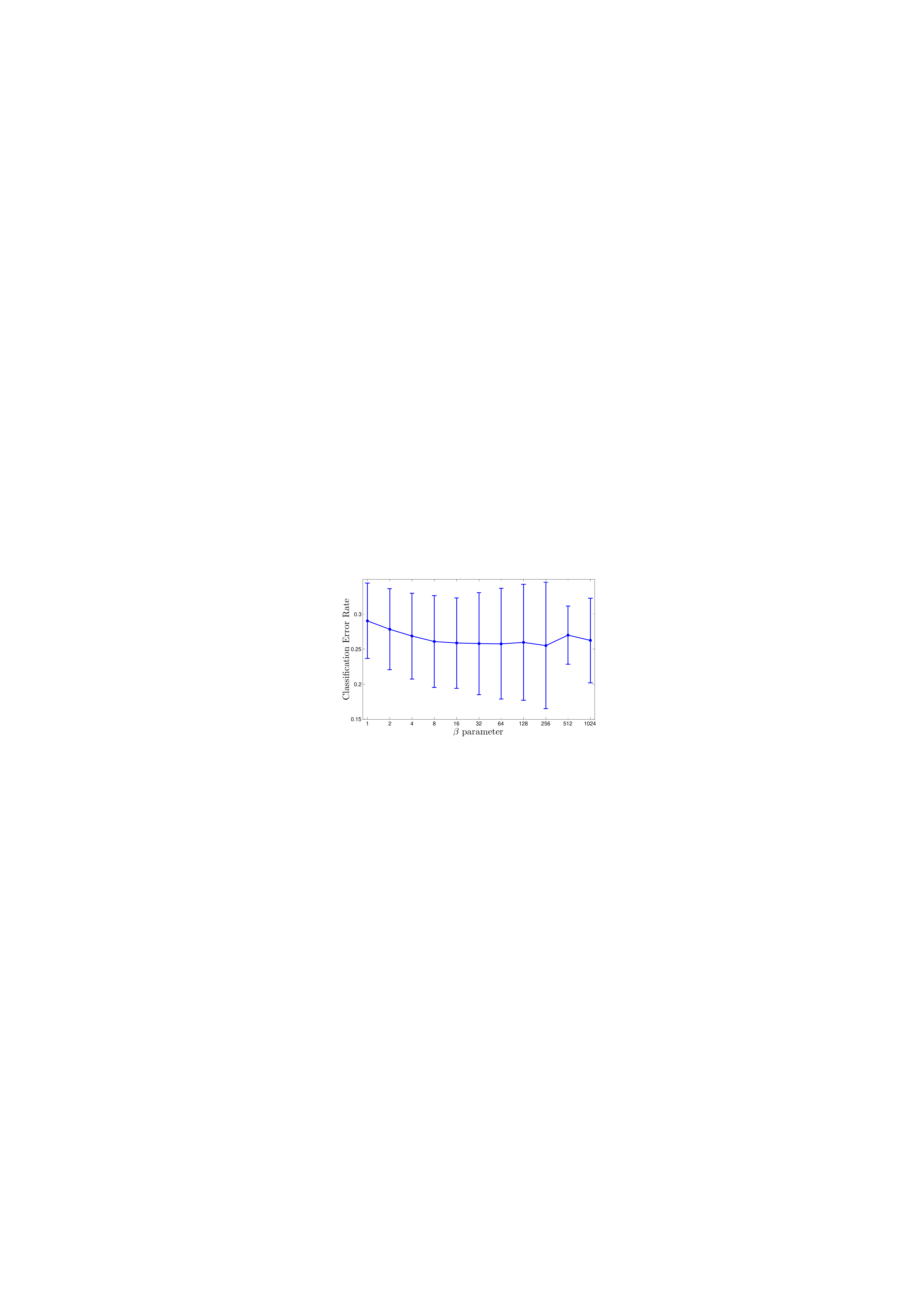}
\caption{Classification error rate and standard deviation with LNR-PCC using different values for $\beta$ parameter, applied to the Semeion Handwritten Digit data set \cite{Semeion1,Semeion2} with $159$ labeled nodes, from which half of them are incorrectly labeled. Each data point is the average of $400$ executions, $4$ on each of $100$ randomly selected labeled subsets.}
\label{fig:beta-semeion}
\end{figure}

The aforementioned parameters $k$ and $\Delta_v$ from LNR-PCC were inherited from its previous versions \cite{Breve2012TKDE,Breve2010IJCNN,Breve2012SBRN}, in which they were extensively studied. LNR-PCC also introduces the new parameter $\beta$. In this paper, we fixed $\beta=10$, as explained in Section \ref{sec:ResetRule}. Two different scenarios from the experiments above were selected to show how the $\beta$ parameter affects the classification accuracy. Figures \ref{fig:beta-g241c} and \ref{fig:beta-semeion} show the classification error rate and standard deviation when the LNR-PCC method is applied to the g241c and the Semeion Handwritten Digit data sets, respectively. Notice that $\beta=1$ is equivalent to not applying the reset rule. In the first scenario (Figure \ref{fig:beta-g241c}), it is clear that the $\beta$ parameter is important to decrease classification error and that it has an optimal value, beyond which the classification error starts to increase again. In the second scenario (Figure \ref{fig:beta-semeion}), there is a larger range of optimal $\beta$ values, but it is still clear that the reset rule is important because $\beta=1$ produces the worst result.

\section{Conclusions}
\label{sec:Conclusions}

In this paper we have proposed a new particle competition and cooperation method for semi-supervised classification in the presence of label noise. Particles walk through the graph generated from the data set. Each particle cooperates to other particles of the same label and competes against particles of dierent labels to classify unlabeled samples. The new algorithm is specifically designed to address the problem of noise label by discovering and re-labeling them, employing novel graph construction rules and new particle dynamics. These built-in features lead to an increased robustness to noise label, preventing noise propagation at a large extent and, therefore, achieving better classification accuracy. Unlike other methods, which requires separate steps for filtering label noise and classifying unlabeled nodes, the proposed method performs the classification of unlabeled data and reclassification of labeled data together in a unique process.

The improvements over the original particle competition and cooperation approach were developed mostly to address the phenomenon that we call \emph{territory switching}, where two or more teams of particles almost completely move to territories that belongs to another class, leaving their own territory to be taken by enemies as well. The new graph construction steps reduces the connectivity of the network, thus preventing particles from taking long trips and keeping them around their home nodes. The individual distance tables also keep particles closer to their home node, as they slightly increases the nominal distance of teammates' home nodes, effect that is also enhanced by the new graph construction steps. Finally, the reset rule brings particles back to home periodically, so that even if the territory switch occasionally occurs, it will not ruin the classification.

Computer simulations were performed using some artificial and real-world data sets with increasing amount of label noise. The experimental results indicate that the proposed model is robust to the presence of label noise. In the comparison to other representative graph-based semi-supervised methods, including previous particle competition and cooperation models, the proposed method presents better classification accuracy in most of the analyzed scenarios.

\section*{Acknowledgment}

The authors would like to thank the S\~{a}o Paulo State Research Foundation (FAPESP) and the National Counsel of Technological and Scientific Development (CNPq) for the financial support.

\section*{References}

\bibliography{Neurocomputing2014}

\end{document}